\documentclass[letterpaper, 10 pt, conference]{ieeeconf}
\usepackage{graphicx} 

\usepackage{subcaption}
\usepackage{colonequals}
\usepackage{amsmath} 
\usepackage{float}
\usepackage{booktabs}
\usepackage{amsfonts}
\usepackage{amssymb}
\usepackage{xcolor}
\usepackage{siunitx}
\usepackage[font=small,labelfont=bf]{caption}
\DeclareMathOperator*{\argmax}{arg\,max}

\usepackage[ruled]{algorithm2e}
\makeatletter
\newcommand{\algorithmstyle}[1]{\renewcommand{\algocf@style}{#1}}
\newcommand{\removelatexerror}{\let\@latex@error\@gobble}
\makeatother

\IEEEoverridecommandlockouts                              

\title{\LARGE \bf
Robot eye-hand coordination learning by watching human demonstrations: a task function approximation approach
}

\author{Jun Jin$^{1}$, Laura Petrich$^{1}$, Masood Dehghan$^{1}$, Zichen Zhang$^{1}$, Martin Jagersand$^{1}$
\thanks{$^{1}$Authors are with Department of Computing Science,
        University of Alberta, Edmonton AB., Canada, T6G 2E8
        {\tt\small jjin5, laurapetrich, masood1, vincent.zhang, martin.jagersand @ualberta.ca}}%
}

\begin{document}

\maketitle
\thispagestyle{empty}
\pagestyle{empty}

\begin{abstract}
We present a robot eye-hand coordination learning method that can directly learn visual task specification by watching human demonstrations. Task specification is represented as a task function, which is learned using inverse reinforcement learning(IRL\cite{Abbeel2004}) by inferring a reward model from state transitions. The learned reward model is then used as continuous feedbacks in an uncalibrated visual servoing(UVS\cite{Jagersand1997}) controller designed for the execution phase. Our proposed method can directly learn from raw videos, which removes the need for hand-engineered task specification. Benefiting from the use of a traditional UVS controller, the training on real robot only happens at initial Jacobian estimation which takes an average of 4-7 seconds for a new task. Besides, the learned policy is independent from a particular robot, thus has the potential of fast adapting to other robot platforms. Various experiments were designed to show that, for a task with certain DOFs, our method can adapt to task/environment changes in target positions, backgrounds, illuminations, and occlusions.

\end{abstract}

\section{Introduction}\label{sec:intro}
We address four problems in robot eye-hand coordination learning by watching. \textbf{1)} How to learn task specification from raw videos of human demonstration? \textbf{2)} How to directly train a visuomotor policy on a real robot with minimal hardware wear-out cost, while not using any simulators? \textbf{3)} How is the learned policy's generalization ability under different task settings (e.g., occlusions, illumination changes)? \textbf{4)} How to make the learned policy independent of this robot and be fast adaptive to a new robot platform?

Learning by watching human demonstrations provides a more intuitive way for task teaching. Recently proposed learning based methods (e.g., end-to-end~\cite{Levine2016},  IRL~\cite{Finn2016}, meta-learning~\cite{yu2018one}) provide strong generalization ability in different task settings, but training either needs simulators~\cite{james2017transferring} or on real robot~\cite{liu2018imitation}~\cite{xie2018few}, in which case, a certain time of training on this particular robot are needed, thus hardware wear-out and safety issues arise.
\begin{figure}
	\begin{center}
		\includegraphics[width=0.48\textwidth]{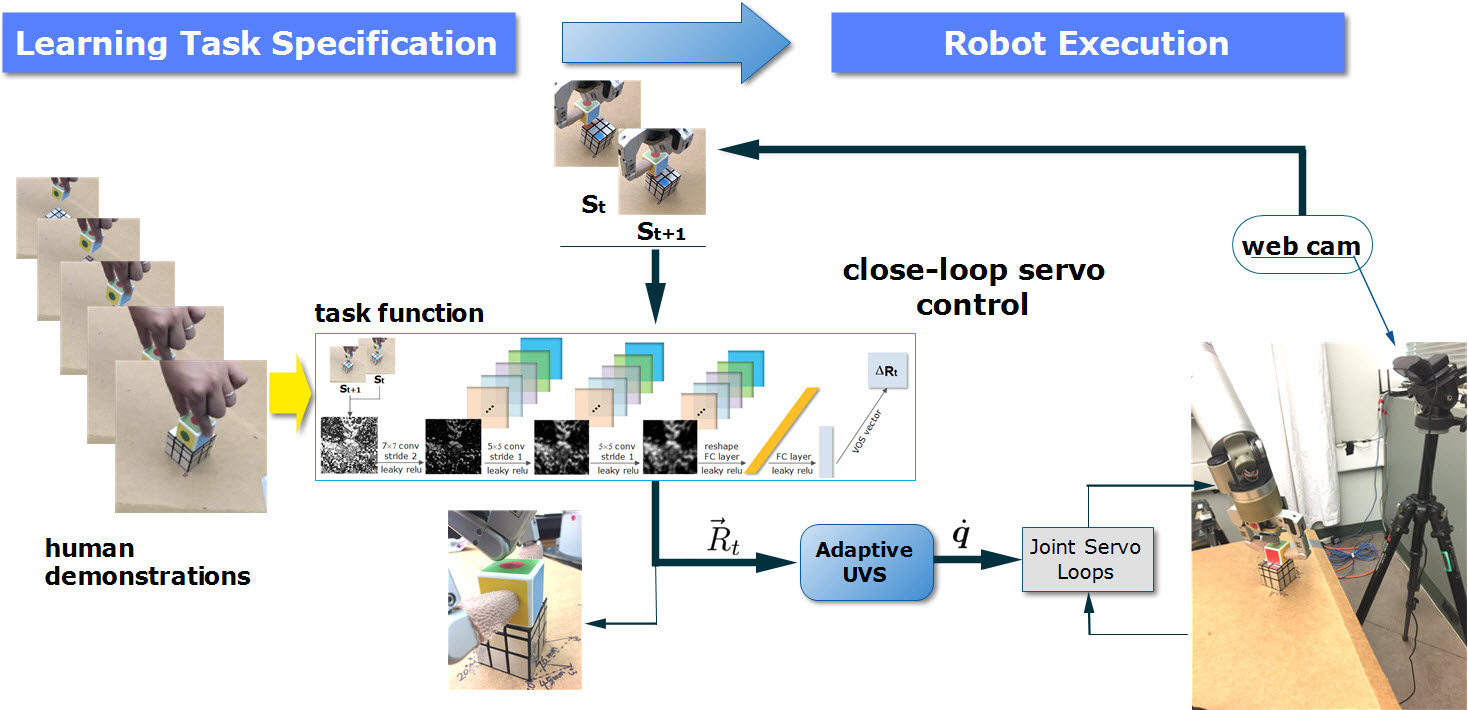} 
		\caption{Our proposed method firstly learns a task function from human demonstration videos. The learned task function receives real-time video streams and outputs a \textit{reward vector} $\mathbf{R_{t}}$, which is used as continuous feedbacks in a close-loop UVS controller to guide robot motions in the execution phase.}
		\label{fig:design_overview}
	\end{center}
\end{figure}
\begin{figure*}
	\centering
	\includegraphics[width=0.98\textwidth]{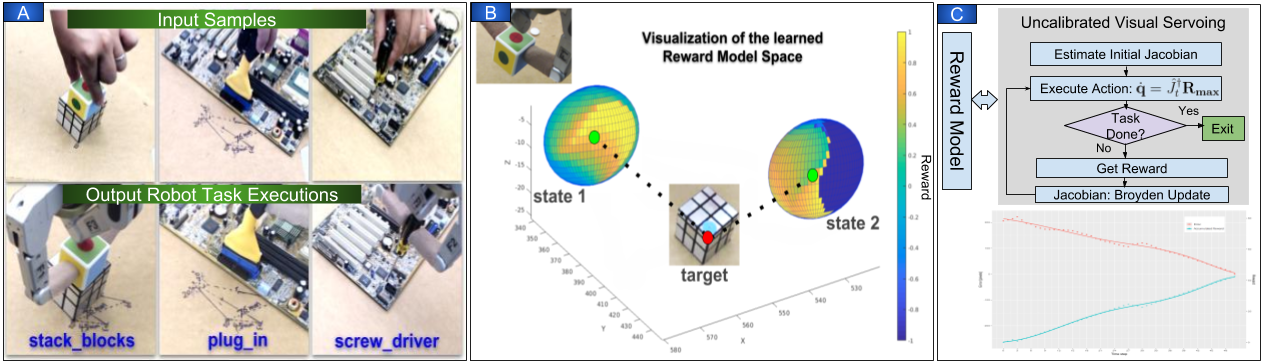}
	\caption{\textbf{A: Task definition.} Given: videos of human demonstrations (above row). Question: Can robot do the same task? How about changing environments(occlusions, illumination changes)? From left to right, it represents a task with 1) regular geometric shape, 2) complex backgrounds and 3) small visual signatures respectively.\textbf{B: Visualization of the learned task function of stacking\_blocks task.} The colored sphere represents the resulting reward when object moves from the center point (green dot) towards any direction on this sphere surface. The red dot defines the target. Left sphere and right sphere show results on different positions. Results show the region of directions going towards the target will return a \textbf{maximum} reward. \textbf{C: Execution phase} Up: The learned task function (reward model) acts as continuous feedback in an uncalibrated visual servoing controller.Down: An execution curve shows a proportional change pattern that, while the accumulated reward is increasing, error (in pixels) is minimized to zero. Errors are measured tediously by logging each step image and marking the pixel distance between object and target by hand.}
	\label{fig:net_design}
\end{figure*}

Compared to learning based approaches, traditional control methods (e.g., visual servoing ~\cite{agin1977servoing,Chaumette2006}) have the advantage of both data and control efficiency~\cite{corke1994high}, good interpretability, almost no hardware wear-out issue, and can be fast adaptive to any practical robot platforms. However, problems regarding cumbersome task specification~\cite{Hager2000}, tracking~\cite{Gridseth2016} and robustness under variances~\cite{bateux2018training}, impede its real-world applications.

The above-mentioned two approaches can be viewed in one unified framework, which is inspired by human eye-hand coordination. Human infants learn eye-hand coordination by first watching demonstrations, and then fulfilling the task in an exploratory manner.~\cite{rao2006cognition}. It is worth mentioning that 1) at a cognitive level, the learning process associates movement patterns with outcomes~\cite{rao2006cognition}, and 2) from a control perspective, the human vision system acts as feedbacks~\cite{Pehoski2006object} in a closed loop motion control manner.

How to relate visually observed outcomes to a motion pattern? In traditional visual servoing, this is more about understanding how tasks are specified and represented. Hager~\cite{Hager2000} and Dodds~\cite{Dodds1999a} proposed a programmable task specification method where an error function (a.k.a., task functions) is composed by setting various combinations of basic geometric constraints (e.g. point-to-point)~\cite{Dodds1999}. This task function represents visual outcomes as a vector. Later, a Jacobian~\cite{Jagersand1997} that relates this vector and motion changes (e.g., joint velocity) is calculated. As we mentioned before, constructing this task function also needs robust tracking on image features.

Is it possible to directly learn a task function from raw image sequence and then use the task function in a traditional controller? We propose a method (Fig. \ref{fig:design_overview}) that learns a task function from raw video inputs based on Inverse Reinforcement Learning (IRL, ~\cite{Abbeel2004}). The \textit{task function} utilizes a \textit{reward vector} to measures the motion outcome observed in image space. It is subsequently used as real-time feedback in a traditional closed-loop visual servoing controller with a minimal (4-7 seconds) of real robot online training. Major contributions are:
\begin{itemize}
	\item Hand-engineered task specification in visual servoing is removed by directly learning from raw video demonstration;
	\item The use of a traditional control method makes directly training on real robot with minimal cost become possible, without using any simulators. It also provides fast adaptive ability to a different robot other than the trained one.
	\item Experimental design shows that for certain degrees of freedom (DOF) tasks, our method can generalize to variations in target positions, backgrounds, illuminations, and occlusions without prior retraining.
	\item Task interpretability in both the training and execution phase are provided by the learned \textit{task function};
\end{itemize}

\section{Methods}
\label{sec:InMaxEnt IRL}
\subsection{Learning task function from raw videos of human demo}
We propose \textbf{Incremental Maximum Entropy IRL (\textit{InMaxEnt IRL})} to learn the task function from demonstrations. Unlike most IRLs, our method learns from state transitions~\cite{fu2017learning,liu2018imitation,Sermanet2016} which has the potential of bringing a better generalization ability~\cite{fu2017learning}. Since it learns on state level, state changes can be incrementally stacked with increased number of demonstrations. For simplicity, we use one demonstration in this section to derive the basics of \textit{InMaxEnt IRL}.

\begin{figure*}
	\centering
	\includegraphics[width=0.9\textwidth]{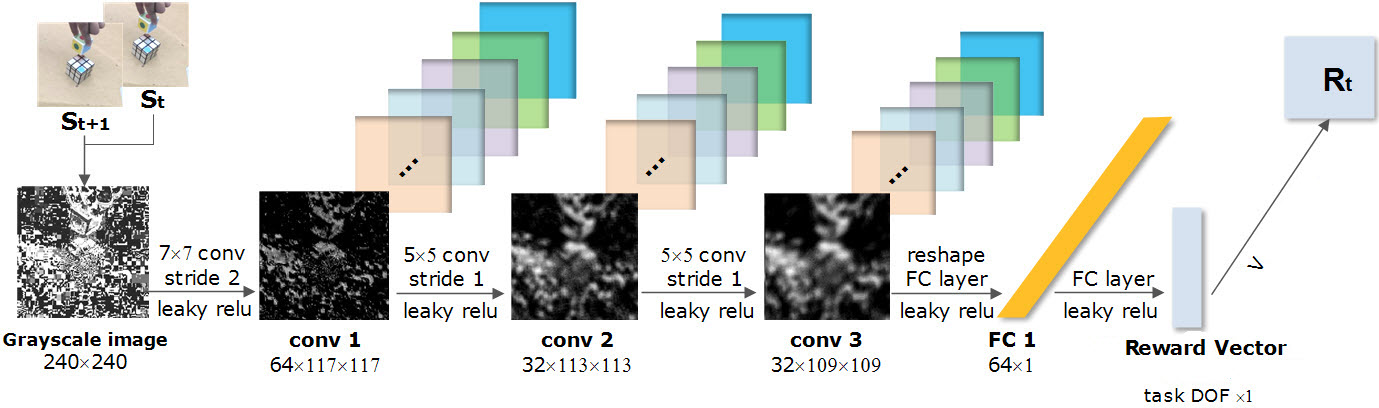}
	\caption{A five-layer neural network is designed to represent \textit{Task Function T}. It receives state change $d{s_{t}}^{+}$ computed by modular subtraction between $s_{t+1}$ and $s_{t}$, while outputs a \textit{Reward Vector} which measures the outcome when moving from $s_{t}$ to $s_{t+1}$. For the purpose of maximum entropy problem formation, \textit{Reward Vector} is then converted to a scalar \textit{Reward} $R_{t}$ by by a dot product with vector $\mathbf{v}$ (defined below), in which case, the \textit{Task Function T} is a \textit{reward function}.}
	\label{fig:net_design}
\end{figure*}

\subsubsection{Terms and Definitions} 
Suppose we have an expert demonstration $\tau_{i}$. Let $s_{t}$ denotes the raw image state sampled at time $t$,and $\tau_{i} = \{s_{0},...,s_{n+1}\}$. Let's define the state change from $s_{t}$ to $s_{t+1}$ as $ds_{t}^{+}$, while the inverse direction from $s_{t+1}$ to $s_{t}$ as $ds_{t}^{-}$. In our experiments, $ds_{t}^{+}$ and $ds_{t}^{-}$ are simply the result of modular subtraction between two images.

The outcome of the state changes is measured using a \textit{Reward vector} : $\mathbf{R_{t}}\in \mathbb{R}^{d}$, where $d$ is the task DOF. For simplicity, we assume each element in $\mathbf{R_{t}}$ is within range $[-1,1]$. \textbf{Our goal is to estimate a \textit{task function}} $T$:
\begin{equation}
\mathbf{R_{t}} = T(ds_{{t}}|\boldsymbol{\theta})
\end{equation} 
, where $ds_{t}$ defines a state change (e.g., $ds_{t}^{+}$ or $ds_{t}^{-}$), and $\boldsymbol{\theta}$ are parameters to learn. In this paper, a neural network (Fig. \ref{fig:net_design}) is used to represent the task function $T$.

In order to have a scalar reward in IRL problem formation, the vector reward $\mathbf{R_{t}}$ is then converted to a scalar $R_{t} \in [-1,1]$ by a dot product with $\mathbf{v}=\frac{1}{d}[1,...,1]^{\top}$.

Let's further define $\mathbf{R_{t}^{+}}$, $R_{t}^{+}$ as the observed state transition case and $\mathbf{R_{t}^{-}}$, $R_{t}^{-}$ as the inverse direction case. They share the same task function with parameters $\boldsymbol{\theta}$ to estimate.
\subsubsection{Generic action ${a_{t}}$}
A generic action ${a_{t}^{+}}$ defines any actions can could deterministically cause state transition from $s_{t}$ to $s_{t+1}$, whereas, ${a_{t}^{-}}$ defines state change from $s_{t+1}$ to $s_{t}$. So a generic action is independent of a specific robot. The mapping from generic actions to particular actions (e.g., joint velocity, torques) is done by a traditional controller in this paper. Global optimal control methods can also be used (e.g., RL\cite{Peters2008}).

\subsubsection{Boltzmann Distribution with Human Factor}
A Boltzmann Distribution is commonly used in Maximum Entropy based IRL~\cite{Ziebart2008}, which measures expert's preference over selection of trajectories. We borrow the same idea to measure expert's preference in selecting actions at $s_{t}$. Also, we argue that an expert is NOT selecting among all possible actions but selecting among high impact actions according to his/her confidence during demonstration. 

For example, if an expert is pretty sure about what actions are optimal at $s_{t}$, he is selecting only from a small set of candidate actions. On the other hand, if he/she is not sure about what actions are `good', the expert will have a hard time making decisions since the candidate action pool becomes large. We measure this behavior using a Boltzmann Distribution with Human Factor. This human factor refers to his/her confidence $\alpha$ during demonstration. The confidence level $\alpha$ will have a resulting variance $\sigma _{0}$, where high confidence level corresponds to low $\sigma _{0}$, vice versa.

Assume in state $s_{t}$, expert's action $a_{t}^{+}$ is selected from a set of possible actions $\mathcal{A}_{t}=$\{$a_{t1}, a_{t2}, ..., a_{tj}, ...$\}, where each action will return a reward $R_{tj}$. In order to emphasize high impact actions in this selection pool, we cast the possibility of $a_{tj}$ appearing in the pool $\mathcal{A}_{t}$ as: $P(R_{tj}) = \mathcal{N}(R_{t}^{+}, \sigma_{0} ^{2})$. $R_{t}^{+}$ is the expert action's reward. So low reward actions will have lower probability appearing in this pool. We call this human factor prior. 

As a result, we represent the probability of selecting action $a_{t}$ at $s_{t}$ as:
\begin{equation}
\label{eq:ha1}
P(a_{t}|s_{t})=\frac{1}{\mathcal{Z}_{t}}exp(R_{t})P(R_{t})
\end{equation} 
, where $a_{t}$ defines a generic action (e.g, $a_{t}^{+}$ or $a_{t}^{-}$) $\mathcal{Z}_{t}=\int_{-1}^{1} exp(R_{tj})P(R_{tj})dR_{tj}$, is the partition function and difficult to estimate.

The role of human factor $\sigma _{0}$ casts how stronger the expert assumption is in IRL. It will determine how much effort should we invest in optimization. Obviously, if an expert is very confident in his/her demonstrations, we should invest greater efforts in optimization since we know it's \textbf{promising}. On the contrary, if an expert is very diffident in demonstration, we should invest less effort in optimization since the expert assumption is \textbf{questionable} now. The effect of using $\sigma _{0}$ will be further discussed later.

So for observed and the inverse sequence, the probability of selecting an observed an expert action $a_{t}^{+}$ and the inverse $a_{t}^{-}$ at $s_{t}$ is:
\begin{equation}
\label{eq:positiveP}
\begin{aligned}
P(a_{t}^{+}|s_{t})=\frac{1}{\mathcal{Z}_{t}}exp(R_{t}^{+})P(R_{t}^{+})\\
P(a_{t}^{-}|s_{t})=\frac{1}{\mathcal{Z}_{t}}exp(R_{t}^{-})P(R_{t}^{-})
 \end{aligned}
\end{equation} 
\subsubsection{Problem Formulation}
Given an expert trajectory $\tau_{i}$, we can formulate the problem by maximizing the likelihood of an observed state sequence $\{s_{0}, s_{1},...,s_{n+1}\}$ while minimizing the possibility of a negative sequence $\{s_{n+1},...,s_{1},s_{0}\}$:
\begin{equation}
\label{eq:cost}
\boldsymbol{\theta} ^{*}  = \argmax_{\boldsymbol{\theta}} \frac{P(s_{0}, s_{1},...,s_{n+1})}{P(s_{n+1},...,s_{1},s_{0})} 
\end{equation} 
By applying the property of MDP, we have $P(s_{0},...,s_{n+1})=P(s_{0}) \prod_{t=0}^{n}P(a_{t}^{+}|s_{t})$. Unless prior knowledge is provided, $P(s_{0})$, the initial prior, can be dropped in optimization. The same rule applies to negative sequence, and eqn. (\ref{eq:cost}) can be written as:
\begin{equation}
\label{eq:cost2}
\boldsymbol{\theta} ^{*}= \argmax_{\boldsymbol{\theta}}\prod_{t=0}^{n}\frac{P(a_{t}^{+}|s_{t})}{P(a_{t+1}^{-}|s_{t+1})}
\end{equation} 
The above function is then expanded by adding the \textit{negative log-posterior}. Combined with eq. (1), it's now a function of parameters $\boldsymbol{\theta}$ with input samples $ds_{t}^{+}$ and $ds_{t}^{-}$. 

Computing the partition function $\mathcal{Z}_{t}$ is challenging. One possible solution is to use importance sampling with the assumption that $P(R_{tj}) \sim \mathcal{N}(R_{t}^{+}, \sigma_{0} ^{2})$, and then construct a generator network to approximate $\mathcal{Z}_{t}$. G plays the role as: given a reward $R_{t}$ at state $s_{t}$, what's the corresponding state change $ds_{t}$. More accurate results could be obtained but the computation cost is expensive. We present a solution using \textit{Boltzmann Factor} as a good trade-off between accuracy and computational cost.

\subsubsection{Simplified Assumption: Boltzmann Factor}
Is it possible to obliterate the partition function $\mathcal{Z}_{t}$ from numerator and denominator in eq. (5), since $P(a_{t}^{+}|s_{t})$ and $P(a_{t}^{-}|s_{t})$ share the same partition function $\mathcal{Z}_{t}$ at state $s_{t}$?

We borrow the concept of \textit{Boltzmann Factor} to measure the relative entropy change between $ds_{t}^{+}$ and $ds_{t}^{-}$:
\begin{equation}
\begin{aligned}
    \boldsymbol{\beta_{t}}& =\frac{P(a_{t}^{+}|s_{t})}{P(a_{t}^{-}|s_{t})}\\
    & =exp(R_{t}^{+} - R_{t}^{-} + \frac{(R_{t}^{+} - R_{t}^{-})^{2}}{2 \sigma_{0}^{2}}).
\end{aligned}
\end{equation}
Now the partition function $\mathcal{Z}_{t}$ is eliminated. Though not accurate, we can further assume that, the initial state transitions $P(a_{0}^{+}|s_{0})$ and $P(a_{n-1}^{-}|s_{n+1})$ in the observed and negative sequence respectively, can be dropped out in optimization. Now eq. (\ref{eq:cost2}) can be rewritten as:
\begin{equation}
\label{eq:cost_b}
\boldsymbol{\theta} ^{*}= \argmax_{\boldsymbol{\theta}} \prod_{t=1}^{n} \boldsymbol{\beta_{t}}
\end{equation} 
By applying the \textit{log-posterior}, we have:
\begin{equation}
\label{eq:cost_f}
\begin{split}
	\boldsymbol{\theta} ^{*}= \argmax_{\boldsymbol{\theta}} \sum_{t=1}^{n}R_{t}^{+} - R_{t}^{-} + \frac{(R_{t}^{2} - R_{t}^{-})^{2}}{2 \sigma_{0}^{2}}
\end{split}
\end{equation}
Since $R_{t}$ is bounded to [-1,1], it's trivial to get the upper bound of the average cost value using $\sigma_{0}^{2}$ as:
\begin{equation}
\label{eq:upper_bound}
upper\_bound = 2(1 + 1/\sigma_{0}^{2})
\end{equation}

Let's continue our discussion on human factor $\sigma_{0}^{2}$ in section A(3). Now the effect is more obvious. For higher confident expert demonstrations, the variance $\sigma_{0}^{2}$ will be smaller, thus have a higher upper bound. The learning process will push more towards this upper bound. For lower confidence demonstrations, vice versa. This is all about how stronger our expert assumption is in IRL.

This simplified version of \textit{InMaxEnt IRL} using stochastic gradient ascend (Algorithm 1) is used in later experiments.

\begingroup
\removelatexerror
\begin{algorithm*}[H]
	\SetAlgoLined
	\KwIn{Expert demonstrations \{$\tau_{1},...,\tau_{m}$\}, task DOF number d, confidence level $\alpha$}
	\KwResult{Optimal weights $\boldsymbol{\theta} ^{*}$ of Task Function T}
	\textcolor[rgb]{0.14,0.36,0.73}{\textbf{Prepare State Change Samples $\mathcal{D}{s}^{+}$, $\mathcal{D}{s^{-}}$}}\\
	\For{i = 1:m}{
		\For{t=1:sample size of $\tau_{i}$}{
			Compute $d{s_{t}^{+}}$, $d{s_{t}}^{-}$\\
			Append samples: $\mathcal{D}{s}^{+} \leftarrow d{s_{t}^{+}}$, $\mathcal{D}{s^{-}} \leftarrow d{s_{t}}^{-}$\\		
		}
	}
	Compute $\sigma_{0}$ using $\alpha$, construct $\mathbf{v}$ using d\\
	\textcolor[rgb]{0.14,0.36,0.73}{\textbf{Shuffle $\mathcal{D}{s}^{+}$, $\mathcal{D}{s^{-}}$; Initialize $\boldsymbol{\theta} ^{0}$}}\\
	\For{n=1:N}{
		\For{t=1:sample size}{
			\textcolor[rgb]{0.14,0.36,0.73}{\textbf{Forward pass}}\\
			${R_{t}^{+}}=T(ds_{t}^{+}, \boldsymbol{\theta})\cdot \mathbf{v}$, ${ R_{t}^{-}}=T(ds_{t}^{-}, \boldsymbol{\theta})\cdot \mathbf{v}$\\
			$ll(\boldsymbol{\theta})=R_{t}^{+} - R_{t}^{-} + \frac{(R_{t}^{+} - R_{t}^{-})^{2}}{2 \sigma_{0}^{2}}$\\
			$err1  =\frac{\partial ll(\boldsymbol{\theta})}{\partial(R_{t}^{+})} \cdot \mathbf{v} $\\ 
			$err2 =\frac{\partial ll(\boldsymbol{\theta})}{\partial(R_{t}^{-})} \cdot \mathbf{v}$\\ 
			$grad1 = T.backProp(ds_{t}^{+},\boldsymbol{\theta} ^{n}, err1)$\\
			$grad2 = T.backProp(ds_{t}^{-},\boldsymbol{\theta} ^{n}, err2)$\\
			\textcolor[rgb]{0.14,0.36,0.73}{\textbf{Gradient ascend update}}\\
			$\boldsymbol{\theta} ^{n+1}=updateWeights(\boldsymbol{\theta} ^{n}, grad1+grad2)$
		}
    }
	\caption{InMaxEnt IRL  {\small (using Boltzmann Factor)} }
\end{algorithm*}
\endgroup

\subsection{Execution: map to a real robot using UVS}
Now we have learned a reward model as the task function, uncalibrated visual servoing (UVS~\cite{Jagersand1997}) is used here to directly approximate an affine mapping from task space to robot joint space. As shown in Fig. 2C, UVS has four steps in a closed-loop control manner: 1) estimate an initial Jacobian by driving random motions and observing changes in task function. This is the only online training process on robot and cost an average of 4-7 seconds depending on hwardware speed. 2) Calculate action (joint velocity) using this Jacobian and execute this action. 3) Observe new changes in task function. 4) Broyden update the Jacobian. 

Since we want an action that results in a maximum reward, which can be defined as $\mathbf{R_{max}} = [1,...,1]^{\top} \in \mathbb{R}^{d}$, where d is the task DOF. Action is computed by:
\begin{equation}
\label{eq:control_law}
\dot{\mathbf{q}} = \hat{J}_{t}^{\dagger}\small \mathbf{R_{max}} \normalsize
\end{equation}
, where $\hat{J}_{t}^{\dagger}$ is the estimated Jacobian.

The biggest challenge is how to handle accumulated drifting error caused by continuous Broyden updates. In practice, we set a threshold vector $\mathbf{R_{thres}}$ to estimate ${\hat{J}_{t+1}^{\dagger}}$'s singularity proximity and perform ${\hat{J}_{0}}$ re-calibration if necessary.

 \begin{figure}
	\begin{center}
		\includegraphics[width=0.4\textwidth]{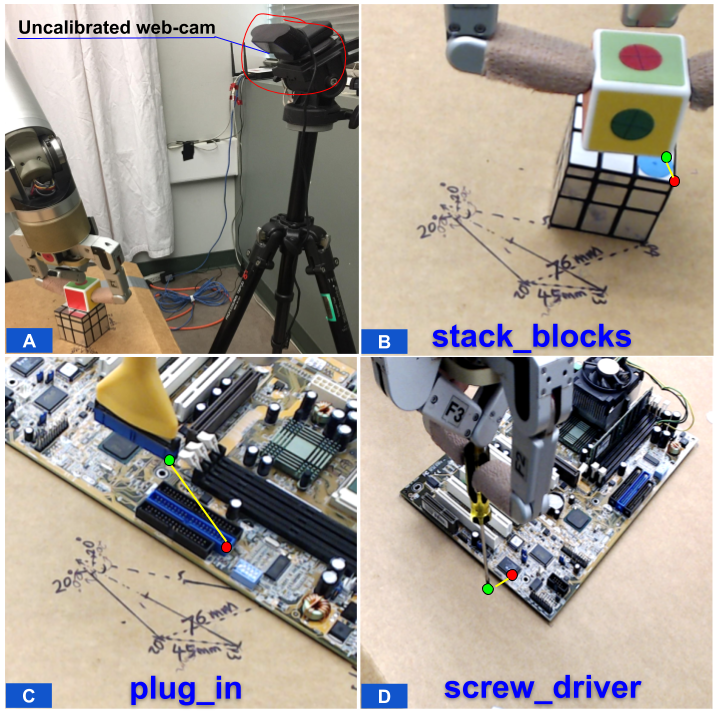} 
		\caption{\textbf{A}: Experimental setup. A low-cost uncalibrated webcam is used to record human demonstrations and guide robot motions. \textbf{B,C and D}: measurement of error using pixel distance between current position(green dot) to target(red dot). A threshold of 20 pixels ($\approx7mm$) in a $580\times580$ image is used to determine a trial's success. Examples of successful and failure trials are shown in Fig. 5.}
		\label{fig:different_tasks} 
	\end{center}
\end{figure}
\begin{figure}
	\begin{center}
		\includegraphics[width=0.48\textwidth]{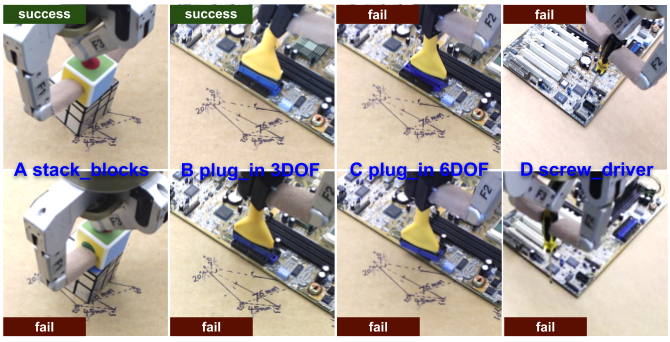} 
		\caption{\textbf{Examples of successful and failure trials in each of the four tasks}. A: stack\_blocks task has a successful rate of \textbf{70\%}. B: Plug\_in 3DOF task is \textbf{60\%}. The other two tasks all failed. The Plug\_in 6DOF task failed since the learned task function is coarse and can't handle high DOF in the final UVS execution phase. The Screw\_driver task failed since image changes between states are very small which cause larger errors in the learned task function.}
		\label{fig:different_tasks} 
	\end{center}
\end{figure}
\section{Experiments} \label{sec:experiments}
Detailed experimental setup and evaluation metric is shown in Fig. 4. Our experimental objective aims to answer the following three questions:
(1) What kind of task is our method capable of and what of task fails? (2) How does generalization differ under variances with tasks and environments? (3) How does performance change with varying number of demonstrations?

\subsubsection{Capability with Different Tasks}
We aim to evaluate four categories(Fig. \ref{fig:different_tasks}) of fine manipulation tasks: (1) a 3DOF task with regular geometric shapes ``stack blocks''; (2) a 3DOF task with complex backgrounds ``plug in a socket on circuit board''; (3) a same ``plug in'' task with 6DOF. (4) a 3DOF task with small visual signatures ``pointing with the tip of a screw driver''. All tasks were trained on a moderate PC (one GTX 1080Ti GPU) using 11 human demonstrations.

Each task includes 10 trials. After each trial, visual error was manually measured (Fig. 4). Results are shown in Table \ref{difftasks}. Examples of success and fail trials of each task can be found in Fig. 5. Learning curves in training the task function, which show how cost is maximized are drawn in Fig. 6.

\begin{table}
	\centering
	\caption{Evaluation results on different tasks. Training on robots happens only in initial jacobian estimation. avg is 7 seconds which is minimum.}\label{difftasks}
	\begin{tabular}{llllll} 
		\toprule
		{\textbf{Task}}&  {\textbf{Training (only images)}} & {\textbf{Successes}} & {\textbf{Mean error}}  \\ 
		\midrule
		stack\_blocks                                                                                   & 7.6 min       & 7/10                                                                                       & 6.3$\pm$2.2                                            \\
		plug\_in                                                                                         & 9.8 min       & 6/10                                                                                          & 6.3$\pm$1.6                                                   \\
		screw\_driver                                                                                    & 13.3 min      & 0/10                                                                                           &  -                                                  \\
		plug\_in 6DOF                                                                                   & 11.84 min     & 0/10                                                                                            &  -                                                  \\
		\midrule
	\end{tabular}
\end{table}

According to results showed in Table \ref{difftasks} and Fig. 6, the proposed method performs moderately well in tasks with regular geometric shapes and complex backgrounds, but fails in small visual change conditions and 6DOF tasks. In the screw\_driver task, image changes are mostly caused by image noise instead of caused by actual robot motions. This also results in a longer training time. For the 6DOF task, as shown in Fig. 2B, the learned task function is still coarse and the adaptive UVS controller is a local method that's difficult to control with a coarse model.

\begin{figure}
	\begin{center}
		\includegraphics[width=0.4\textwidth]{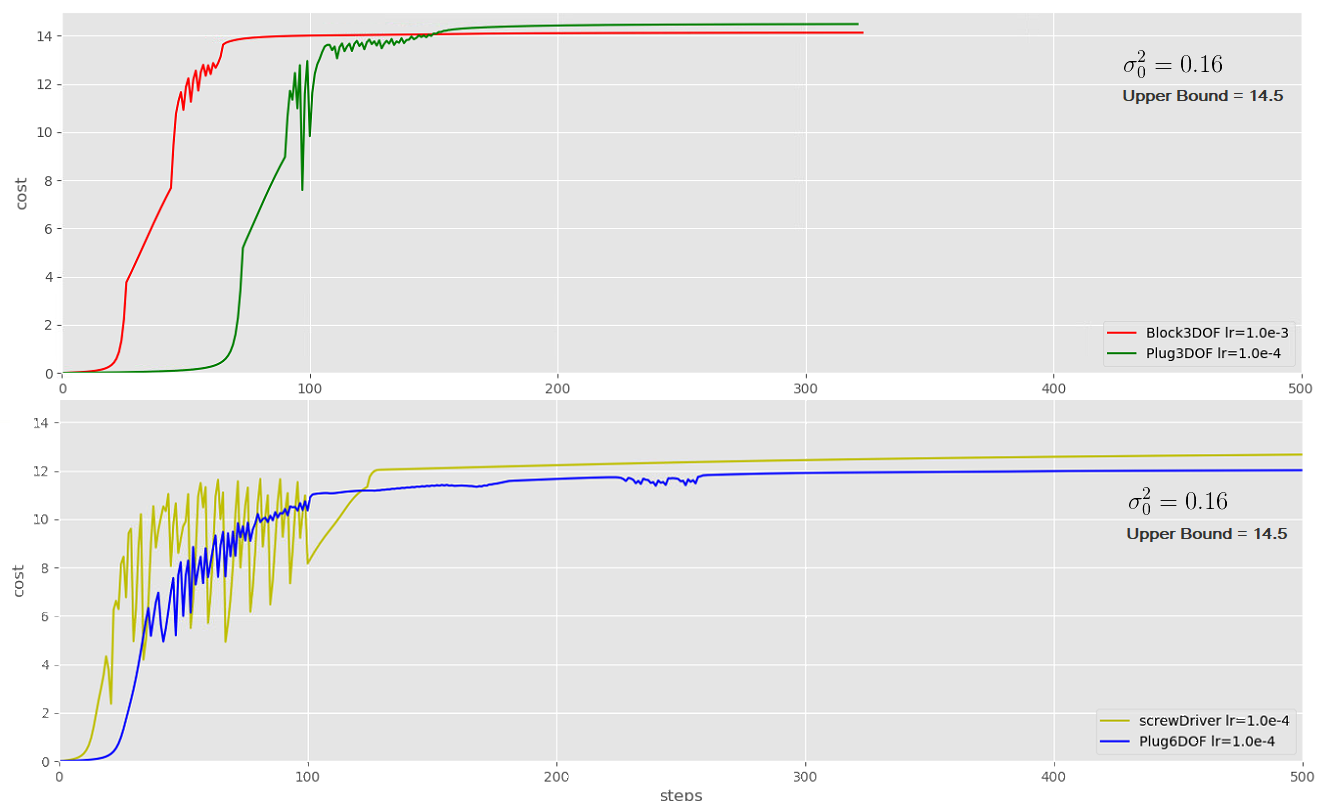} 
		\caption{Learning curves of all the 4 tasks in training the task function. All have the same variance $\sigma _{0}^{2}=0.16$ with corresponding upper bound 14.5 calculated in eq. (9). \textbf{Up:} Curves of stack\_blocks task (red) and plug\_in task (green). These two tasks have equal success rate and training pushes the cost very near to the upper bound. The resulting task function is more accurate. \textbf{Down:} Curves of plug\_in 6DOF (blue) task and screw\_driver (yellow) task. These two tasks all failed. It shows difficulty in training, and can't not push the cost near the upper bound. The resulting task function is more coarse thus causing failures.}
		\label{fig:different_tasks} 
	\end{center}
\end{figure}
\begin{figure*}
	\centering
	\includegraphics[width=1.0\textwidth]{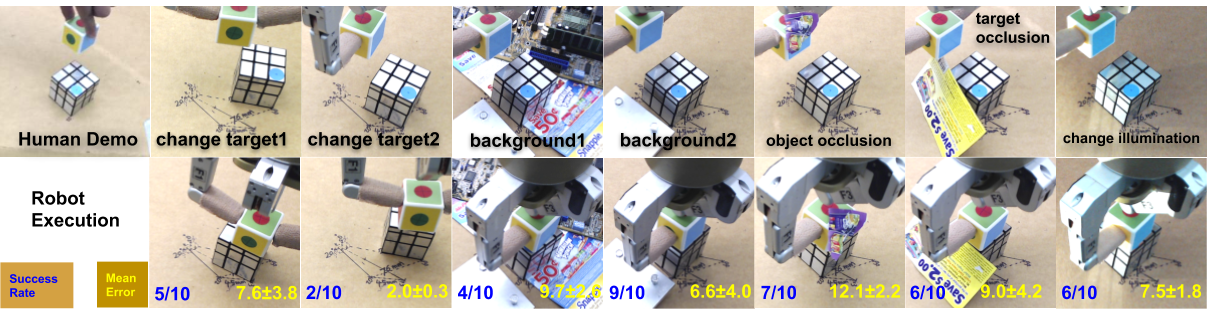}
	\caption{Evaluation setup under different task/environment settings. \small Up row: initial settings; Down row: robot execution results, details are shown in Table \ref{tb_conditions}. Results show that it can generalize well under moderately changed target positions and backgrounds, occlusions and illumination changes. \normalsize }
	\label{fig:diff_conditions}
\end{figure*}
\begin{table}
	\begin{center}
	\caption{Evaluation results of generalization ability. Results show that it can generalize well under moderately changed target positions and backgrounds, occlusions and illumination changes.}\label{tb_conditions}
	\begin{tabular}{llllll} 
		\toprule
		{\textbf{Variant settings}}&  {\textbf{Avg. steps}} & {\textbf{Successes}} & {\textbf{Error (Pixel)}}  \\ 
		\midrule
		(1) change\_target1& 23 & 5/10 & \hspace{0.13cm}7.6$\pm$3.8  \\
	(2) change\_target2& 22 & 2/10 & \hspace{0.13cm}2.0$\pm$0.3  \\
	(3) background1& 45 & 4/10 & \hspace{0.13cm}9.7$\pm$2.6  \\
	(4) background2& 24 & 9/10 & \hspace{0.13cm}6.6$\pm$4.0  \\
	(5) object\_occlusion& 26 & 7/10 & 12.1$\pm$2.2  \\
	(6) target\_occlusion& 36 & 6/10 & \hspace{0.13cm}9.0$\pm$4.2  \\
	(7) change\_illumination& 30 & 6/10 & \hspace{0.13cm}7.5$\pm$1.8  \\
		\midrule
	\end{tabular}
	\end{center}
\end{table}
\subsubsection{Performance under Variances with Tasks and Environments}
We tested the following settings (Fig. \ref{fig:diff_conditions}): (1) change\_target1: The target block is translated 45mm long and rotated with \ang{20}; (2) change\_target2: The target block is translated 75mm long and rotated with \ang{40}; (3) background1: Background is changed to a super messy one; (4) background2: Background is changed by adding an extra object; (5) object\_occlusion: One face of the small block is occluded; (6) target\_occlusion: One face of the target block is occluded; (7) change\_illumination: An extra light source is placed directly opposite to the scene. Each of the seven setting had 10 trials. After each trial, visual error was manually measured following the same rule as stated before.

Results (Table \ref{tb_conditions}) show that it can generalize well under moderately changed target positions and backgrounds, occlusions and illumination changes. Since the state images are further processed using modular subtraction, it's not surprising that environment changes does not affect the performance as much, but the method performs poorly with large target or background changes.

\subsubsection{Performance as Sample Size Increases}
We are also interested in performance evaluation with varing numbers of human demonstrations. Using the same task ``plug\_in'', we trained using 1, 5, and 11 human demonstrations, respectively. For each trained model, 10 trials were conducted and visual error was again used as a performance measure. Results are shown in Table \ref{table_s}.
\begin{table}
	\centering
	\caption{Evaluation results using different number of human demonstrations. Results show that the performance improves when using more demonstrations.}\label{table_s}
	\begin{tabular}{llll} 
		\toprule
		{\textbf{Demo Num}} & \textbf{Training (only images)} & {\textbf{Successes}} & {\textbf{Mean error}}  \\ 
		\midrule
		1                                                      & 2.0 min                & 1/10                                            & 10.8$\pm$0                                           \\
		5                                                      & 5.0 min                & 2/10                                            & 16.3$\pm$0.5                                          \\
		11                                                     & 9.8 min                & 6/10                                            & \hspace{0.13cm}6.3$\pm$1.6                                       \\
		\midrule
	\end{tabular}
\end{table}
\subsubsection{Discussion on failures}
Failure trials mainly come from two aspects: (i) the accuracy of task function, especially in cases with changing target position with patterns unseen in training samples. That's also the reason why increasing human demonstration numbers will improve its performance. (ii) the Jacobian estimation in UVS control since quality of $\mathbf{\hat{J}_{0}}$ has a large effect on the control convergence. Jacobian estimation error mostly comes from the strategy of switching between Broyden update and ${\hat{J}_{0}}$ re-calibration, which is similar to the \textit{exploration vs. exploitation} problem in Reinforcement Learning.
	
\section{Previous Works}
Our work was inspired by research advances in visual servoing, including defining a task function using geometric constraints~\cite{Hager2000,Dodds1999a,Gridseth2016,quintero2017flexible,dehghan2018online}, direct visual servoing (DVS~\cite{silveira2012direct}) method to remove the tracking challenge, and increasing generalization ability in visual servoing~\cite{bateux2018training} using deep neural networks. DVS is very similiar to our approach, however, it relies on a planar assumption and inconvenient task specification process.

Human demonstrations in place of manual task specification with end-to-end learning have been proposed to address these challenges~\cite{Levine2016}. IRL seeks to derive a \textit{reward function} from observable actions and is closely related to learning from demonstration (also known as imitation learning or apprenticeship learning)~\cite{Abbeel2004}. This learned reward function is synonymous with the visual servoing error function, notwithstanding the scalar output of the reward function, whereas the error function outputs a vector with dimensionality determined by task DOF. Maximum entropy IRL was proposed to manage the problem of sub-optimal demonstrations \cite{Ziebart2008}, and Wulfmeier et al. represented the reward function using neural networks to handle non-linearity~\cite{Wulfmeier2015}. A challenge still exists in the estimation of the partition function $Z$, since it must solve the entire Markov Decision Process (MDP); this can be computationally expensive and unfeasible to generalize in a large action space or under unknown system dynamics. Finn et al. proposed an iterative solution using importance sampling to approximate a soft optimal policy\cite{Finn2016} and recent works revealed the connection between IRL and generative adversarial networks (GAN)~\cite{Ho2016,finn2016connection}. Limitations of learning based approaches can be found in section \ref{sec:intro}.

There are also other approaches on task specification learning, which are trying to build a task tree at the semantic level~\cite{Dillmann2004}\cite{Yang2015}\cite{Xiong2016} and essential in practice to provide high level task programming.

\section{Conclusions}\label{sec:conclusions}
We present a robot eye-hand coordination learning method that can directly learn task specification by watching raw demonstration videos, and map to joint velocity control using an adaptive UVS controller. \textit{InMaxEnt IRL} is proposed to infer a task function from human demonstration videos. The use of a traditional controller enables efficient training on real robot with minimal hardware wear-out cost (4-7 seconds). It's also independent of a specific robot and in theory, provides fast adaptive ability on other robots.

\textbf{Limitations and future directions:} The major limitation of our method comes from its \textit{local optimality}, not only derived from the reward based task function model, but also from the affine mapping in UVS control. Although it has this limitation, it provides the possibility to use global optimal control methods (e.g RL\cite{Peters2008}) training directly on real-world robotic systems. Another limitation is the \textit{accuracy of task function}, which is the main reason why higher DOF tasks failed. Besides, the assumption of Boltzman Factor can't work in non-prehensile tasks as $P(a_{t}^{-}|s_{t}$ is zero.

Future work could look at improving our simple network design to increase accuracy, for example through utilizing deeper networks, combing geometric invariant learning, and/or gaze selection\cite{Johansson2001}. Furthermore, with regards to task interpretability, it's still unclear how the \textit{differential reward vector} is related to a high DOF task, and visual ambiguity from the single-camera view exists. The generalization to multiple view cameras could prove fruitful for future research.

\section*{Acknowledgement}
The first author is grateful for receiving funding on his living expenses and tuition from China Scholarship Council (CSC) and the University of Alberta during his study.

\addtolength{\textheight}{-2cm}   





\bibliographystyle{IEEEtran}
\bibliography{IEEEabrv,IEEEexample}

\end{document}